\documentclass[11pt]{article}

\usepackage[preprint]{acl}

\usepackage{times}
\usepackage{latexsym}

\usepackage[T1]{fontenc}

\usepackage[utf8]{inputenc}

\usepackage{microtype}

\usepackage{inconsolata}

\usepackage{graphicx}
\usepackage{amsmath,amssymb,amsfonts}

\usepackage{array}

\usepackage{graphicx}

\usepackage[table]{xcolor}
\usepackage{colortbl}
\usepackage{booktabs}
\usepackage{arydshln}

\usepackage{pifont}

\usepackage{amssymb}
\usepackage{graphicx}
\usepackage{tcolorbox}
\tcbuselibrary{skins, breakable}
\usepackage{subcaption}

\usepackage{multirow}       
\usepackage{dashrule}
\definecolor{myblue}{RGB}{198,222,239}
\newcommand{\cb}[1]{\cellcolor{myblue}#1}
\usepackage{fontawesome5}
\definecolor{other flat1}{RGB}{175,0,75}

\definecolor{other flat1}{RGB}{175,0,75}
\newcommand{\hi}[1]{\textbf{\textcolor{other flat1}{#1}}}
\def\BibTeX{{\rm B\kern-.05em{\sc i\kern-.025em b}\kern-.08em
    T\kern-.1667em\lower.7ex\hbox{E}\kern-.125emX}}

\definecolor{neg green}{RGB}{0,120,90}
\newcommand{\lo}[1]{\textbf{\textcolor{neg green}{#1}}}

\usepackage[ruled,vlined,linesnumbered]{algorithm2e}

\usepackage{xcolor}
\newcommand{\algcmt}[1]{\textcolor{gray}{#1}}

\DontPrintSemicolon
\SetKwComment{Comment}{$\triangleright$\ }{} 

\title{When Models Judge Themselves: Unsupervised Self-Evolution for Multimodal Reasoning}

\author{
 \textbf{Zhengxian Wu\textsuperscript{1,2}$^\dagger$},
 \textbf{Kai Shi\textsuperscript{1}$^{\dagger\ddagger}$},
 \textbf{Chuanrui Zhang\textsuperscript{3}},
 \textbf{Zirui Liao\textsuperscript{2}},
 \textbf{Jun Yang\textsuperscript{1}$^*$},
 \textbf{Ni Yang\textsuperscript{1}},
 \textbf{Qiuying Peng\textsuperscript{1}},
\\
 \textbf{Luyuan Zhang \textsuperscript{2}},
 \textbf{Hangrui Xu\textsuperscript{4}},
 \textbf{Tianhuang Su\textsuperscript{1}},
 \textbf{Zhenyu Yang\textsuperscript{1}},
 \textbf{Haonan Lu\textsuperscript{1}},
 \textbf{Haoqian Wang\textsuperscript{2}\thanks{
   $\dagger$ Equal contribution \quad
   $\ddagger$ Project leader \quad
   Co-first authors: zx-wu24@mails.tsinghua.edu.cn, shikai@oppo.com \\
   $^*$ Corresponding authors: yangjun2@oppo.com, wanghaoqian@tsinghua.edu
 }},
\\
 \textsuperscript{1}OPPO AI Center,
 \textsuperscript{2}Tsinghua University,
 \textsuperscript{3}Nanyang Technological University,
 \textsuperscript{4}Hefei University of Technology
}

\begin{document}
\maketitle
\begin{abstract}
Recent progress in multimodal large language models has led to strong performance on reasoning tasks, but these improvements largely rely on high-quality annotated data or teacher-model distillation, both of which are costly and difficult to scale.
To address this, we propose an unsupervised self-evolution training framework for multimodal reasoning that achieves stable performance improvements without using human-annotated answers or external reward models. 
For each input, we sample multiple reasoning trajectories and jointly model their within group structure.
We use the Actor’s self-consistency signal as a training prior, and introduce a bounded Judge based modulation to continuously reweight trajectories of different quality.
We further model the modulated scores as a group level distribution and convert absolute scores into relative advantages within each group, enabling more robust policy updates. 
Trained with Group Relative Policy Optimization (GRPO) on unlabeled data, our method consistently improves reasoning performance and generalization on five mathematical reasoning benchmarks, offering a scalable path toward self-evolving multimodal models.
The code is available at https://github.com/OPPO-Mente-Lab/LLM-Self-Judge.
\end{abstract}

\section{Introduction}
In recent years, multimodal large language models (MLLMs) have demonstrated remarkable progress in vision–language reasoning tasks.
These models have achieved impressive performance on a wide range of benchmarks, including visual mathematical reasoning\cite{visionr1}, chart understanding\cite{chartmuseum}, and complex scene inference\cite{seeingBT}.


However, much of this progress still relies on high-quality training data and strong supervision signals\cite{Li2025PerceptionRT}. 
Such supervision usually comes from carefully annotated answers and reasoning traces\cite{Safaei2025FilterIF}, or from stronger models or evaluators trained on expensive preference data, whose capabilities are then transferred to the target model through distillation\cite{visionr1}.
At the same time, obtaining such supervision at scale is becoming increasingly costly. 
High-quality annotated data is growing more scarce, and the capability of existing evaluators is also approaching its practical limit\cite{Tao2025LimitedPD}.
Motivated by this challenge, recent studies have begun to explore self-evolving post-training for multimodal models. 
The goal is to reduce reliance on human annotation and external supervision, and instead use unlabeled data to automatically construct training signals for further improving reasoning ability\cite{Vision-zero}.

\begin{figure*}[t]
    \centering
    \includegraphics[width=\textwidth]{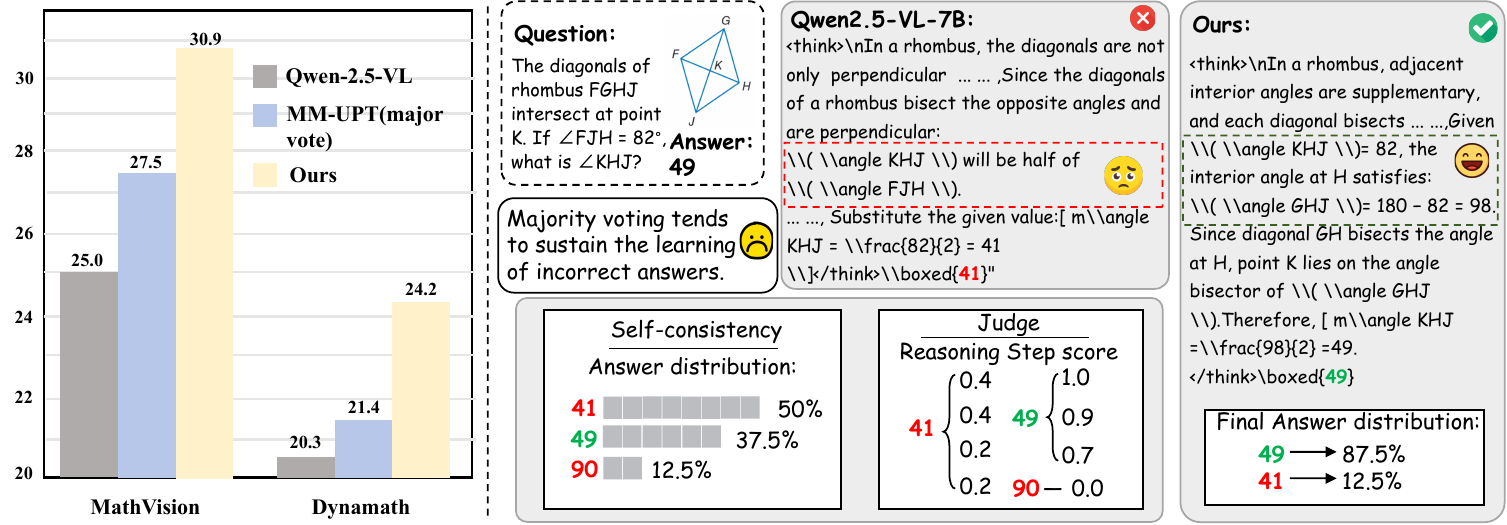}
    \caption{Limitation of majority voting in unsupervised self-evolution.Right: An example where the most frequent answer is incorrect. Majority voting reinforces this dominant error, while our method favors higher-quality reasoning paths through Judge modulation.Left: Results on MathVision\cite{mathvision} and DynaMath\cite{Dynamath} show that our approach consistently outperforms majority-voting-based self-training.}
    \label{fig1}
    \vspace{-0.4cm}
\end{figure*}

The main challenge of self-evolving post-training is the lack of reliable supervision, which makes the training signals noisy and biased.
Applying reinforcement learning on top of such signals further increases the risk of gradient fluctuation and training instability.
Existing approaches\cite{MM-UPT,EvoLMM}  often use model-generated intermediate results or pseudo-labels as training signals. A common strategy is to sample multiple responses and measure their consistency.
Some recent work\cite{Zhou2025EvolvingLM} further introduces diversity or novelty signals to encourage exploration. 
In practice, this strategy provides a “bootstrapped” approximation of stable supervision: it reduces the noise of a single sample and aligns the training objective with output patterns that are relatively consistent under the current policy distribution.
Nevertheless, as illustrated in Fig~\ref{fig1}, high consistency does not necessarily imply high quality; it may instead reflect systematic biases of the model, which can be amplified during long-term training and suppress effective exploration. 
Moreover, the training signal fails to capture fine-grained differences between candidates and can further trigger response-length collapse. 
As training proceeds, rewards often concentrate quickly on a few dominant modes, causing optimization to saturate early and pushing the policy toward a low-entropy output distribution.

In light of these limitations, we argue that stable unsupervised self-evolution should strike a balance between robustness and effectiveness.
Motivated by this insight, we propose a self-evolving training framework.
Specifically, we instantiate two roles from a single multimodal model: an Actor and a Judge.
Given an input, the Actor samples multiple reasoning trajectories, forming the model’s current self-consistency distribution.
The Judge evaluates each trajectory and maps its score to a bounded and continuously differentiable modulation signal, which calibrates and reshapes the Actor’s initial self-consistency distribution.
On the optimization side, we further construct training rewards in a group-wise, distributional manner.
For multiple trajectories generated from the same input, we apply an energy-based normalization to compare them relatively, converting absolute scores that are not directly comparable across samples into within-group relative advantages.
In this way, training no longer simply amplifies early dominant modes.
Instead, our framework can distinguish fine-grained quality differences among reasoning trajectories for the same input and adjust the model’s output distribution accordingly.
This encourages the optimization objective to better reflect the relative quality among candidate trajectories, leading to more effective improvements in reasoning ability.

We conduct a series of experiments to analyze the limitations of existing paradigms for modeling training signals.
Based on these observations, we further propose and validate a collaborative modeling paradigm.
This paradigm leads to more stable training behavior across benchmarks, for example reflected by healthier entropy trajectories and reduced response-length collapse.
It also delivers more effective performance improvements.
For instance, on MathVision\cite{mathvision}, our unsupervised post-training achieves up to a +5.9 absolute improvement in accuracy (30.9\% vs. 25.0\%).
Importantly, the entire training pipeline does not rely on ground-truth labels, additional metadata, or any external reward model at any stage.
In summary, our main contributions are as follows:
\begin{enumerate}
    \item We propose a new framework for unsupervised post-training of large multimodal models, enabling sustained self-improvement without any external supervision.
    \item Through extensive empirical analysis, we identify common failure modes in unsupervised self-evolution and mitigate them by modeling and optimizing the within-input relative structure among candidate solutions.
    \item We evaluate our method on multiple mathematical reasoning benchmarks and observe accuracy improvements after multiple iterations under different training data settings.
\end{enumerate}

\section{Related work}
\subsection{Multi-modal Reasoning}
Motivated by the success of verifiable rewards in LLM reasoning, recent studies\cite{VLM-R1} have begun to explore post-training and R1-style reinforcement learning in multimodal settings.
Instead of relying on subjective human preferences, these methods\cite{R1-Onevision,visionr1} derive reward signals from objectively verifiable signals, enabling more stable reasoning optimization.
Later work\cite{R3V,LLAVA-Critic-R1} integrates reflection into training by using structured reflection steps or learning an explicit critic for evaluation.
NaturalReasoning\cite{Yuan2025NaturalReasoningRI} proposes a method for constructing large-scale reasoning data from real-world corpora. 
Building on this line of work, NaturalThoughts\cite{Li2025NaturalThoughtsSA} studies which teacher-generated reasoning traces are the most useful for distillation. 
R2-MultiOmnia\cite{ranaldi-etal-2025-r2} presents a self-training framework for multilingual multimodal reasoning. 
Despite these advances, effective reasoning post-training still relies on high-quality training signals or stronger teacher models.
\subsection{Self-Evolving In Large Language Models}
Unsupervised self-evolution has been explored to some extent in large language models\cite{CAN_LARGE_REASONING_MODELS_SELF-TRAIN?}.
A core idea is that, even without ground-truth answers, test-time scaling strategies (e.g., majority voting) can provide useful relative correctness signals\cite{TTRL,ETTRL}.
Self-Empowering VLMs\cite{Yang2025SelfEmpoweringVA} studies hierarchical understanding in VLMs and shows that the main challenge is not missing taxonomic knowledge, but the difficulty of maintaining cross-level consistency during step-by-step prediction.
Recently, self-evolution has also been extended to multimodal large language models.
MM-UPT\cite{MM-UPT} uses majority voting over multiple sampled answers to form pseudo-rewards, enabling continual improvement on multimodal reasoning data without ground-truth labels.
However, most of these methods use majority voting as the main training signal, which primarily reinforces consistency under the current output distribution. 

\begin{figure*}[t]
    \centering
    \includegraphics[width=\textwidth]{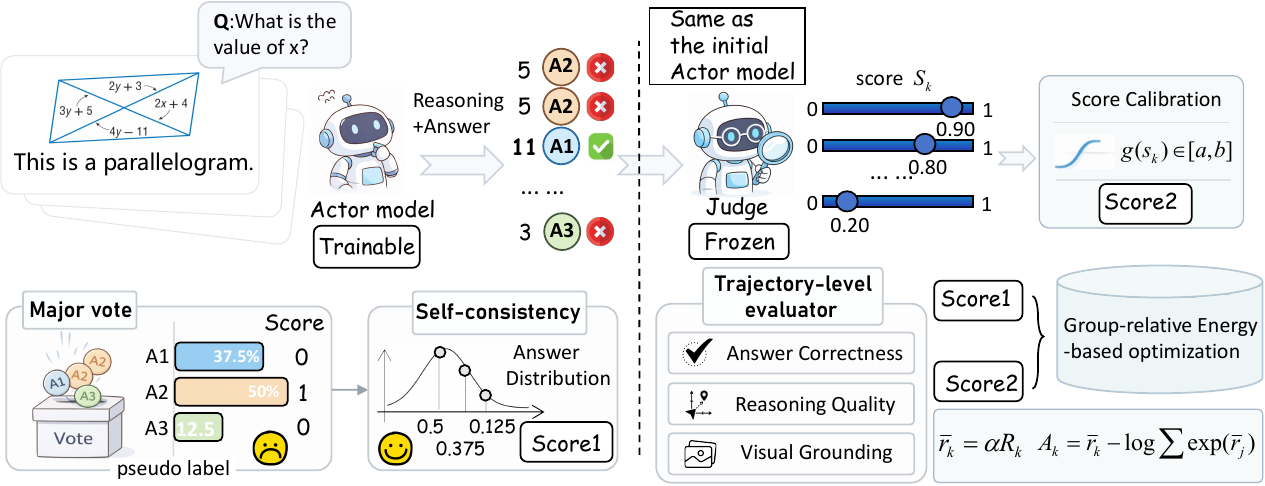}
    \caption{\textbf{Overview of the proposed unsupervised self-evolution framework.}The Actor generates multiple reasoning trajectories for the same input, while a frozen Judge provides bounded score modulation. The final rewards are optimized in a group-wise, distributional manner to enable stable policy updates without external supervision.}
    \label{fig2}
    \vspace{-0.3cm}
\end{figure*}

\section{Method}
As shown in Fig.~\ref{fig2}, we propose an unsupervised self-evolution framework for multimodal large models. 
By jointly modeling multiple reasoning trajectories generated from the same input, our approach enables stable and sustained improvements in reasoning ability.
Specifically, Sec.~\ref{sec:1} constructs a consistency-based initial reward for the Actor from repeated rollouts under the same input. 
Sec.~\ref{sec:2} introduces a Judge to provide a bounded and continuous modulation of this reward.
Finally, Sec.~\ref{sec:3} models the modulated rewards as a group-wise distribution to support more robust policy updates in the unsupervised setting.
\subsection{Consistency-Based Initial Reward for the Actor}
\label{sec:1}
We consider an unsupervised multimodal reasoning sample consisting of an image--question pair 
$x = (I, q)$. 
Given the current policy $\pi_\theta$, we perform $n$ rollouts for the same input $x$, 
resulting in a set of candidate reasoning trajectories:
\begin{equation}
\mathcal{T}(x) = \{\tau_i\}_{i=1}^n, \quad \tau_i \sim \pi_\theta(\cdot \mid x).
\label{eq:trajectory_set}
\end{equation}
Each trajectory $\tau_i$ is associated with a final answer $a_i \in \mathcal{A}$, 
where $\mathcal{A}$ denotes the set of unique answers produced for the input $x$ 
under the current rollouts.
For each answer $a \in \mathcal{A}(x)$, we define its count and the corresponding empirical distribution as:
\begin{equation}
c(a) = \sum_{i=1}^n \mathbb{I}[a_i = a], 
\qquad 
\hat{p}(a) = \frac{c(a)}{n}.
\label{eq:self_consistency_dist}
\end{equation}
We then define the initial reward of each trajectory $\tau_i$ as the empirical frequency of its answer:
\begin{equation}
r_i^{\mathrm{SC}} \triangleq \hat{p}(a_i)
= \frac{1}{n} \sum_{j=1}^n \mathbb{I}[a_j = a_i].
\label{eq:self_consistency_reward}
\end{equation}
Under this formulation, when multiple sampled trajectories agree on the same final answer, 
the corresponding empirical probability $\hat{p}(a)$ becomes larger, and all trajectories associated with 
that answer receive higher rewards accordingly.

\textbf{Consistency-Based Rewards vs. Majority Voting.}
Unlike supervised learning, training signals in unsupervised self-evolution are typically generated by the model itself and therefore inevitably contain noise and bias.
Applying reinforcement learning based optimization on top of such signals often leads to gradient fluctuations and unstable training.
In unsupervised self-evolution, a commonly used paradigm is majority voting, which treats the most frequent answer as the sole training signal.
Formally, it selects the majority answer as:
\begin{equation}
a^{\star}
=
\arg\max_{a \in \mathcal{A}(x)} \hat{p}(a),
\label{eq:majority_answer}
\end{equation}
and assigns a binary reward to each trajectory:
\begin{equation}
r_i^{\mathrm{MV}}
=
\mathbb{I}[a_i = a^{\star}].
\label{eq:majority_reward}
\end{equation}

From the perspective of empirical performance, majority voting is effective in unsupervised self-evolution because it provides a simple denoising mechanism.
By aggregating multiple samples from the same input, it encourages the learning objective to align with outputs that are more consistent under the policy distribution, thereby reducing the randomness of single-sample supervision.
Compared with using raw frequency-based signals, binarized pseudo-labels offer a clearer optimization direction, making it easier for policy updates to obtain noticeable improvements in the early stage.

However, an answer that becomes dominant early in training does not necessarily correspond to a higher-quality reasoning path.
At the same time, the initial answer distribution encodes rich structural information about the model’s output behavior, such as the relative proximity between dominant and secondary modes.
Majority voting discards this information entirely, retaining only the identity of the most frequent answer.
As a result, once an answer becomes dominant at an early stage, the binary reward further amplifies its advantage, driving the policy distribution toward that mode and suppressing exploration of alternative reasoning trajectories.
Over long-term training, this mechanism encourages rapid collapse toward low-entropy, near-deterministic policies.
In contrast, consistency-based rewards preserve the relative strength of the empirical distribution, leading to a smoother training signal and better maintaining effective exploration during optimization.

\subsection{Calibrating Consistency Rewards with a Judge}
\label{sec:2}
The initial reward assigned to the Actor primarily reflects the degree of self-consistency under the current policy, rather than directly measuring the quality or correctness of the underlying reasoning. 
In practice, the model may converge to a pseudo-stable state during training.

To address this issue, we introduce a Judge module that provides a continuous quality signal for each trajectory, serving as a correction to the initial reward. 
Specifically, at the beginning of training, we initialize the Judge as a structurally identical copy of the current Actor policy and keep its parameters fixed throughout training.
The Judge then outputs a raw score for each trajectory by jointly assessing answer correctness, reasoning quality, and visual grounding:
\begin{equation}
s_k = J_\phi(x, \tau_k), \quad s_k \in [0,1].
\label{eq:judge_score}
\end{equation}
Importantly, the Judge score $s_k$ is not used as the final reward directly. 
Instead, it serves as a modulation signal that adjusts the initial reward distribution (see Sec.~\ref{sec:3}). 
To transform the raw Judge score into a stable and controllable modulation signal, we design a calibration function $g(s)$ that satisfies three desiderata: (1) it is continuously differentiable to support stable optimization; (2) it provides appropriate encouragement for high-scoring trajectories and suppression for low-scoring ones; and (3) it is bounded, preventing Judge noise from being amplified in the unsupervised training loop. 
Concretely, we adopt:
\begin{equation}
g(s)
=
1
+
\lambda_{+}\,\sigma\!\Big(\frac{s - t_h}{\tau_h}\Big)
-
\lambda_{-}\,\sigma\!\Big(\frac{t_l - s}{\tau_l}\Big),
\label{eq:judge_calibration}
\end{equation}
where $\sigma(\cdot)$ denotes the sigmoid function, $t_h$ and $t_l$ are the high and low gating thresholds, $\tau_h, \tau_l > 0$ control the smoothness of the gating transitions, and $\lambda_{+}, \lambda_{-} > 0$ determine the maximum magnitude of reward amplification and suppression.

This design incorporates the Judge as a bounded and continuous modulation signal rather than an absolute authority, thereby mitigating pseudo-consistency while avoiding excessive reliance on the Judge’s raw scale in the unsupervised training loop.
More importantly, this joint modeling makes the training signal adaptive.
As the policy distribution evolves, the Judge modulation continuously reshapes the reward signal, preventing optimization from simply locking into the current consensus and enabling ongoing correction during training.

Meanwhile, we also consider a more direct alternative that uses the Judge’s raw score $s_k$ as the reward for optimization.
This choice often leads to instability in an unsupervised closed loop:
since the Judge scores are not comparable in scale across inputs, updates can be dominated by a small number of high-scoring trajectories, causing rapid shifts in the policy distribution.
This shift further amplifies the impact of Judge noise or bias in the training loop, ultimately causing the model to prematurely converge toward the Judge's preference.

\begin{table*}[t]
\centering
\renewcommand{\arraystretch}{1.15}
\resizebox{\textwidth}{!}{
\begin{tabular}{c|c|ccccc|c}
\toprule
\textbf{Training Data} & \textbf{Method} &
\textbf{MathVision} & \textbf{MathVerse} & \textbf{WeMath} &
\textbf{LogicVista} & \textbf{DynaMath} & \textbf{Avg.} \\
\hline

\multicolumn{8}{c}{\textit{Performance on Qwen2.5-VL-7B}} \\
\hline
--- & Qwen2.5-VL-7B & 25.0 & 44.2 & 37.1 & 46.3 & 20.3 & 34.6 \\
\hline

\multicolumn{8}{c}{\textit{Supervised Training \& Teacher Model Distillation}} \\
\hline
\multirow{3}{*}{Mixed Data\faTag} 
& R1-Onevision-7B\cite{R1-Onevision}   & \textbf{29.9} & 46.4 & 35.8 & 46.5 & 21.8 & 36.1 \\
& OpenVLThinker-8B\cite{Deng2025OpenVLThinkerCV}  & 25.9 & 50.3 & \textbf{36.5} & 46.3 & 21.2 & 36.0 \\
& Vision-R1\cite{visionr1}         & 29.4 & \textbf{52.4} & 35.4 & \textbf{49.7} & \textbf{25.2} & \textbf{38.4} \\
\hline

\multicolumn{8}{c}{\textit{Unsupervised self-Evolving}} \\
\hline
CLEVR       & VisionZero\cite{Vision-zero}      & 27.6 & 46.4 & 38.8 & 48.8 & 21.7 & 36.7 \\
ImgEdit     & VisionZero      & 27.4 & 46.8 & 38.5 & 49.1 & 21.3 & 36.6 \\
Multi-Bench & EvoLMM\cite{EvoLMM}          & 25.8 & 44.7 & 37.6 & 46.9 & 21.0 & 35.2 \\
\hline

\multirow{2}{*}{MMR1~\faTag} & +SFT       & 27.3 & 45.0 & 38.3 & 48.1 & 22.9 & 36.3 \\
                             & +RL(GRPO) & 29.3 & 47.4 & 39.3 & 49.4 & 23.3 & 37.7 \\
\hdashline
\multirow{2}{*}{MMR1} &
\cb{MM-UPT\cite{MM-UPT}} & \cb{26.4} & \cb{46.0} & \cb{38.6} & \cb{47.9} & \cb{21.8} & \cb{36.1} \\
& \cb{Ours}
& \cb{\textbf{28.4}} & \cb{\textbf{46.4}} & \cb{\textbf{38.8}}
& \cb{\textbf{48.6}} & \cb{\textbf{23.0}} & \cb{\textbf{37.0}} \\

\hline

\multirow{2}{*}{GeoQA~\faTag} & +SFT       & 27.6 & 45.3 & 38.6 & 47.8 & 22.6 & 36.4 \\
                              & +RL(GRPO) & 28.8 & 47.1 & 39.0 & 49.2 & 23.4 & 37.5 \\
\hdashline
\multirow{2}{*}{GeoQA} &
\cb{MM-UPT} & \cb{27.3} & \cb{45.1} & \cb{38.2} & \cb{47.3} & \cb{21.9} & \cb{36.0} \\
& \cb{Ours}
& \cb{\textbf{28.6}} & \cb{\textbf{46.5}} & \cb{\textbf{38.9}}
& \cb{\textbf{47.9}} & \cb{\textbf{23.2}} & \cb{\textbf{37.0}} \\

\hline

\multirow{2}{*}{Geo3K~\faTag} & +SFT       & 27.8 & 44.7 & 37.9 & 47.2 & 22.1 & 35.9 \\
                              & +RL(GRPO) & 29.1 & 46.9 & 39.1 & 49.6 & 23.8 & 37.7 \\
\hdashline
\multirow{2}{*}{Geo3K} &
\cb{MM-UPT} & \cb{27.5} & \cb{44.0} & \cb{37.4} & \cb{46.9} & \cb{21.4} & \cb{35.4} \\
& \cb{Ours}
& \cb{\textbf{30.9}} & \cb{\textbf{46.8}} & \cb{\textbf{38.7}}
& \cb{\textbf{49.0}} & \cb{\textbf{24.2}} & \cb{\textbf{37.9}} \\

\hline
\end{tabular}}
\caption{\textbf{Main results on multimodal mathematical reasoning benchmarks.}
We report accuracy (\%) on five math benchmarks.
MajorVote corresponds to the MM-UPT method.
\faTag\ denotes supervised training.}
\label{tab1}
\end{table*}

\subsection{Distributional Modeling of the Final Reward}
\label{sec:3}
For the $k$-th trajectory corresponding to the same input $x$, the final reward is defined as:
\begin{equation}
R_k
=
r_k \cdot g(s_k)
-
\lambda_{\mathrm{fmt}}\, \delta_k,
\label{eq:final_reward}
\end{equation}
where $\delta_k \in \{0,1\}$ indicates whether the trajectory violates the predefined output format constraints, and $\lambda_{\mathrm{fmt}} = 0.5$ is the corresponding penalty coefficient.
We adopt Group Relative Policy Optimization (GRPO)\cite{GRPO} to perform relative optimization over candidate trajectories corresponding to the same input. 
For a given input $x$, let the reward vector of its $n$ trajectories be $r(x) = [R_1, \ldots, R_n]$. 
We first apply energy-based scaling to the rewards:
\begin{equation}
\tilde{r}_k = \alpha\, R_k,  
\label{eq:reward_scaling}
\end{equation}
where $\alpha$ is a temperature parameter.
We then define a group-wise log-sum-exp baseline as:
\begin{equation}
b(x) = \log \sum_{j=1}^n \exp(\tilde{r}_j),
\label{eq:lse_baseline}
\end{equation}
The resulting group-relative advantage is computed as:
\begin{equation}
A_k(x) = \tilde{r}_k - b(x).
\label{eq:group_advantage}
\end{equation}
Importantly, this construction implicitly induces a reward-defined target distribution over the candidate set:
\begin{equation}
q_\alpha(\tau_k\mid x)
=
\frac{\exp(\alpha R_k)}
{\sum_{j=1}^{n}\exp(\alpha R_j)}.
\end{equation}
It then follows that:
\begin{equation}
A_k(x) = \log q_\alpha(\tau_k\mid x).
\end{equation}

This shows that the group-relative advantage corresponds to the log-probability of a trajectory under the reward-induced distribution.
Therefore, the policy update can be understood as gradually matching the current policy to this target distribution:
\begin{equation}
\min_{\theta}\;
\mathbb{E}_{x\sim\mathcal{D}}
\Big[
D_{\mathrm{KL}}\big(
q_\alpha(\cdot\mid x)\,\|\,\pi_\theta(\cdot\mid x)
\big)
\Big].
\end{equation}

A more detailed derivation is provided in Appendix~\ref{sec:appendix1}.
By modeling the final scores as a group-wise distribution, policy updates no longer collapse rapidly to a deterministic mapping. 
Instead, the policy is encouraged to gradually shift probability toward better trajectories, while still keeping several reasonable candidates.
Finally, the GRPO objective for policy optimization can be written as:

{\scriptsize
\begin{equation}
\mathcal{J}_{\mathrm{GRPO}}(\theta)
=
\mathbb{E}
\left[
\frac{1}{n}\sum_{k=1}^{n} r_k^{\mathrm{clip}}
-
\beta\,D_{\mathrm{KL}}\!\Big(\pi_\theta(\cdot\mid x)\,\Big\|\,\pi_{\mathrm{ref}}(\cdot\mid x)\Big)
\right],
\end{equation}

\begin{equation}
r_k^{\mathrm{clip}}
=
\min\!\Big(
\gamma_k(\theta)\,A_k,\;
\mathrm{clip}\!\big(\gamma_k(\theta),\,1-\epsilon,\,1+\epsilon\big)\,A_k
\Big),
\end{equation}

\begin{equation}
\gamma_k(\theta)
=
\frac{\pi_\theta(\tau_k\mid x)}{\pi_{\theta_{\mathrm{old}}}(\tau_k\mid x)}.
\end{equation}
}
\noindent
Here, the expectation is taken over training inputs $x \sim \mathcal{D}$ and the corresponding trajectories $\{\tau_k\}_{k=1}^n$ sampled from the behavior policy $\pi_{\theta_{\mathrm{old}}}(\cdot\mid x)$, $A_k$ denotes the group-relative advantage for the $k$-th trajectory under input $x$, $\gamma_k(\theta)$ is the probability ratio between the current policy and the behavior policy, $\epsilon$ is the clipping threshold, and $\beta$ controls the strength of the KL regularization toward the reference policy $\pi_{\mathrm{ref}}$.

Overall, this group-wise distributional modeling shifts the optimization objective from simply pursuing absolute high scores to continuously reallocating probability mass within each trajectory group, leading to more stable policy updates and reducing the self-reinforcement of early dominant modes in the unsupervised loop.
A more detailed analysis is provided in Appendix~\ref{sec:appendix1}.

\begin{table}[t]
\centering
\resizebox{1.0\linewidth}{!}{
\begin{tabular}{lccc}
\toprule
Method & MathVision & DynaMath & Avg. \\
\midrule
Qwen2.5-VL-7B
& 25.0 & 20.3 & 22.65 \\

+ Major Vote
& 27.5 & 21.4 & 24.45$\mathrm{\tiny{\hi{+1.8}}}$ \\

+ Self-Consistency
& 25.2 & 20.5 & 22.85$\mathrm{\tiny{\hi{+0.2}}}$ \\

+ Judge Scoring
& 27.3 & 21.1 & 24.20$\mathrm{\tiny{\hi{+1.6}}}$ \\

+ MV + JS
& 28.4 & 22.7 & 25.55$\mathrm{\tiny{\hi{+2.9}}}$ \\

+ SC + JS
& 30.1 & 23.7 & 26.90$\mathrm{\tiny{\hi{+4.3}}}$ \\

\rowcolor{gray!10}
+ SC + JS (Dist.)
& \textbf{30.9}
& \textbf{24.2}
& \textbf{27.55$\mathrm{\tiny{\hi{+4.9}}}$} \\

\bottomrule
\end{tabular}
}
\caption{Ablation experiments on different modules.}
\label{tab2}
\end{table}

\begin{table}[t]
\centering
\LARGE
\renewcommand{\arraystretch}{1.4}
\setlength{\tabcolsep}{6pt}

\resizebox{1.15\linewidth}{!}{
\begin{tabular}{lcc|lcc}
\toprule
Model / Method & MathVision & DynaMath &
Model / Method & MathVision & DynaMath \\
\midrule

\rowcolor{gray!10}
Qwen2-VL-2B & 12.8 & 6.8 &
Qwen3-VL-8B & 53.0 & 48.2 \\
w/ major vote & 9.6 & 4.2 &
w/ major vote & 51.7 & 46.8 \\
w/ ours & \textbf{16.3} & \textbf{11.4} &
w/ ours & \textbf{53.5} & \textbf{50.3} \\

\midrule

\rowcolor{gray!10}
Qwen2.5-VL-3B & 19.5 & 18.2 &
GLM-4.1V-9B & 47.8 & 38.6 \\
w/ major vote & 21.3 & 19.1 &
w/ major vote & 48.2 & 37.2 \\
w/ ours & \textbf{24.8} & \textbf{21.3} &
w/ ours & \textbf{50.4} & \textbf{40.8} \\

\midrule

\rowcolor{gray!10}
InternVL3-8B & 20.8 & 23.6 &
Qwen2.5-VL-32B & 38.6 & 35.2 \\
w/ major vote & 22.8 & 23.7 &
w/ major vote & 40.2 & 37.1 \\
w/ ours & \textbf{25.6} & \textbf{26.9} &
w/ ours & \textbf{43.4} & \textbf{38.9} \\

\bottomrule
\end{tabular}
}

\caption{Self-evolution performance comparison across different backbone models.}
\label{tab3}
\end{table}

\begin{figure*}[t]
    \includegraphics[width=\textwidth]{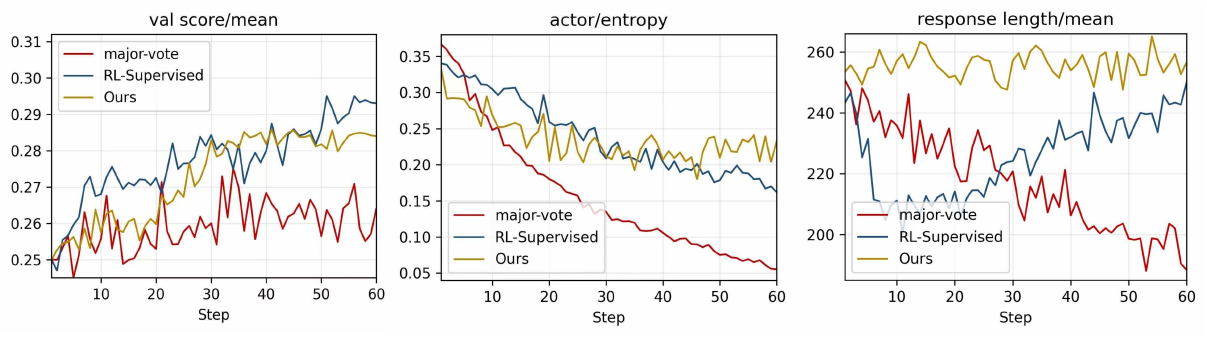}
    \caption{\textbf{Training dynamics on MMR1\cite{mmr1}.}The figure compares majority voting, supervised reinforcement learning, and our method during training, in terms of validation accuracy on MathVision, actor entropy, and average response length.}
    \label{fig3}
\end{figure*}

\begin{figure*}[t]
    \centering
    \includegraphics[width=\textwidth]{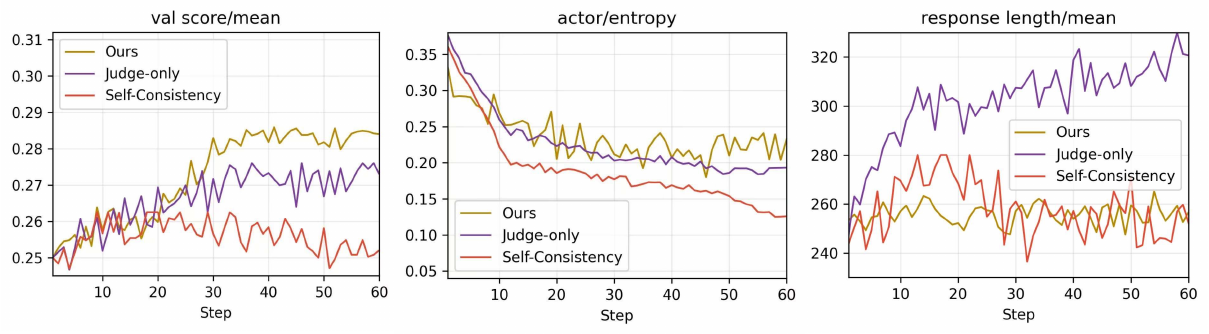}
    \caption{\textbf{Ablation training dynamics on MMR1\cite{mmr1}.}We compare Self-Consistency, Judge-only, and the full method in terms of validation accuracy on MathVision, actor entropy, and average response length during training.}
    \label{fig4}
\end{figure*}

\section{Experiments}
\subsection{Datasets and Training Details}
\label{sec:datasets_training}


\paragraph{Training Data.}
We use Geometry3k\cite{geo3k}, GeoQA\cite{geoqa}, and MMR1\cite{mmr1} as training datasets. 
All experiments are conducted using Qwen2.5-VL-7B-Instruct\cite{qwen2.5-vl} as the backbone.

\paragraph{Evaluation Benchmarks.}
We evaluate our method on several widely used multimodal mathematical reasoning benchmarks, including MathVision\cite{mathvision}, MathVerse\cite{mathvista}, WeMath\cite{we-math}, LogicVista\cite{logicvista}, and DynaMath\cite{Dynamath}, following their standard accuracy protocols. 
We compare our approach against state-of-the-art multimodal unsupervised self-evolving methods, including VisionZero\cite{Vision-zero}, EvoLMM\cite{EvoLMM}, and MM-UPT (major-vote)\cite{MM-UPT}, as well as supervised training schemes such as SFT\cite{SFT-method} and RL-based\cite{RL-method} methods.

\paragraph{Training Setup.}
We perform multimodal unsupervised post-training using the Verl framework\cite{verl}. 
Specifically, both the actor model and the Judge model are initialized from Qwen2.5-VL-7B-Instruct, with the Judge kept frozen while the actor is trained using GRPO\cite{GRPO} for unsupervised reasoning improvement. 
Training is conducted on a single node equipped with $8\times$ NVIDIA A800 GPUs (80GB). 
We set the number of training epochs to 20 and use the AdamW optimizer. 
For the Judge, the sampling temperature is set to 1.0 with top-$p$ sampling of 0.9. 
The reward modulation parameters are set to $\lambda_{+}=\lambda_{-}=0.2$, $t_h=0.95$, $t_l=0.40$, and $\tau_h=\tau_l=1$. 
For distributional reward modeling, the energy-based scaling coefficient is set to $\alpha=1$.
During actor training, each question is rolled out with 8 trajectories. 
The KL-divergence constraint coefficient in GRPO is set to $\beta=0.01$ for training. 
The learning rate is set to $1\times10^{-6}$, with a weight decay of $1\times10^{-2}$ and a gradient norm of 1.0.

\begin{table}[t]
\centering
\resizebox{1.0\linewidth}{!}{
\begin{tabular}{lccc}
\toprule
Method & MathVision & DynaMath & Avg. \\
\midrule

\rowcolor{gray!10}
Qwen2.5-VL-7B
& 0.66 & 0.48 & 0.57 \\

GRPO (w.\ label)
& 0.63 & 0.43 & 0.53$\mathrm{\tiny{\lo{-0.04}}}$ \\

Major Vote
& 0.58 & 0.37 & 0.48$\mathrm{\tiny{\lo{-0.09}}}$ \\

Ours
& 0.64 & 0.43 & 0.54$\mathrm{\tiny{\lo{-0.03}}}$ \\

\bottomrule
\end{tabular}
}
\caption{Comparison of pass@10 across benchmarks.}
\label{tab4}
\end{table}

\begin{table}[t]
\centering
\small
\renewcommand{\arraystretch}{1.05}
\setlength{\tabcolsep}{5pt}
\begin{tabular}{lccc}
\toprule
Model / Method & MathVision & LogicVista & DynaMath \\
\midrule
Vision-R1-7B & 29.4 & 49.7 & 25.2 \\
w/ major vote & 30.0 & 49.4 & 24.6 \\
w/ ours & \textbf{32.6} & \textbf{51.5} & \textbf{26.3} \\
\bottomrule
\end{tabular}
\caption{Self-evolution results on a strong baseline that has already been trained with teacher distillation.}
\label{tab5}
\end{table}

\begin{table}[t]
\centering
\small
\renewcommand{\arraystretch}{1.1}
\setlength{\tabcolsep}{6pt}
\begin{tabular}{lcc}
\toprule
Method & Relative Time & Accuracy \\
\midrule
Supervised GRPO       & 1.0$\times$ & 29.1 \\
MM-UPT (Major Vote)   & 1.0$\times$ & 27.5 \\
VisionZero            & 2.8$\times$ & 27.6 \\
EvoLMM                & 2.2$\times$ & 25.8 \\
\textbf{Ours}         & \textbf{1.4$\times$} & \textbf{30.9} \\
\bottomrule
\end{tabular}
\caption{Comparison of relative training cost and MathVision accuracy across different methods.}
\label{tab6}
\end{table}

\begin{table}[t]
\centering
\small
\renewcommand{\arraystretch}{1.12}
\setlength{\tabcolsep}{5pt}
\begin{tabular}{llcc}
\toprule
Training Setting & Method & ChartQA & MMVP \\
\midrule
Base model & --- & 85.7 & 77.2 \\
\midrule
\multirow{2}{*}{Geo3K}
& w/ MM-UPT & 84.8 & 76.4 \\
& w/ ours   & \textbf{85.4} & \textbf{77.6} \\
\midrule
\multirow{2}{*}{GeoQA}
& w/ MM-UPT & 82.6 & 77.4 \\
& w/ ours   & \textbf{86.3} & \textbf{77.9} \\
\midrule
\multirow{2}{*}{MMR1}
& w/ MM-UPT & 85.4 & 74.6 \\
& w/ ours   & \textbf{87.2} & \textbf{79.8} \\
\bottomrule
\end{tabular}
\caption{Generalization to chart understanding (ChartQA) and general visual reasoning (MMVP).}
\label{tab7}
\end{table}

\subsection{Experimental Results}
\label{sec:experimental_results}

\paragraph{Main Results.}
Table~\ref{tab1} summarizes comparisons between our method and three categories of baselines: (1) the Qwen2.5-VL-7B model without training; (2) state-of-the-art multimodal unsupervised self-evolving methods; and (3) supervised training methods and approaches based on strong-model distillation.
Without relying on any human-annotated answers, our method consistently improves over the original model when trained on all three unsupervised training datasets(MMR1, GeoQA, and Geo3K).
For example, when trained on Geo3K in an unsupervised setting, our method improves the average accuracy from 34.6 to 37.9 (+3.3) across benchmarks. 
The gains are more pronounced on challenging benchmarks. On MathVision, our method achieves an absolute improvement of up to 5.9 points (30.9 vs.\ 25.0).
Compared with existing unsupervised self-evolving methods, our approach consistently outperforms prior work under the same training setting.
Moreover, our method achieves performance comparable to supervised training and strong-model distillation methods, and even surpasses them in some settings.


Figure~\ref{fig3} compares the training dynamics of different strategies on MMR1.
Majority voting rapidly amplifies early dominant answers, leading to a sharp reduction in policy entropy. 
In contrast, our method avoids repeatedly reinforcing early dominant patterns and maintains healthier entropy and response-length trajectories than supervised RL during training, resulting in more stable training behavior.
Meanwhile, we evaluate three unsupervised-trained models on two non-mathematical, vision-centric benchmarks: ChartQA (chart understanding)\cite{masry-etal-2022-chartqa} and MMVP (general visual reasoning)\cite{Tong2024EyesWS}.
Table~\ref{tab7} shows that the proposed training paradigm generalizes beyond mathematical reasoning to broader vision-centric multimodal tasks.
\subsection{Ablation Study}
Table~\ref{tab2} reports ablation results for the key components, while Fig.~\ref{fig4} further illustrates their training dynamics. 
When trained on MMR1, using Self-Consistency alone can retain some output diversity, but it leads to limited improvement on MathVision because it cannot reliably distinguish between highly consistent and low-quality trajectories.
In contrast, the Judge-only variant updates the policy based directly on evaluation scores. 
Although this introduces a quality signal, it ignores the Actor’s candidate distribution within each input, making the updates more likely to be dominated by a small number of high-scoring trajectories.
This leads to unstable training behavior and a noticeable increase in response length. 
Overall, our method achieves the best performance by continuously redistributing probability mass and correcting errors within the candidate set for each input, which helps prevent self-reinforcement of early dominant patterns and reduces training instability.

Table~\ref{tab3} shows the results of applying our method to multiple backbone models of different scales\cite{Zhu2025InternVL3EA,Hong2025GLM45VAG,Yang2024Qwen2TR}. 
Our method remains effective on both weaker models and larger models, demonstrating good scalability across model sizes.
As shown in Table~\ref{tab4}, our method consistently outperforms majority voting on pass@10 across benchmarks. 
Even compared with supervised GRPO (54\% vs.\ 53\%), our method still shows more stable pass@10 performance.
As shown in Table~\ref{tab5}, we further apply self-evolution training to a strong baseline that has already been improved by supervised training and teacher distillation. 
The results show that, even on a model already strengthened by teacher distillation, our online self-evolution mechanism can still bring further gains.
Table~\ref{tab6} compares the computational cost of different methods. 
To ensure a fair comparison, we conduct all experiments under the same hardware setting and report results relative to supervised GRPO (training time $\approx$ 10.5 hours). 
Due to the online sampling and scoring process, our method introduces a moderate additional overhead (1.4$\times$ relative time). 
Further analysis and experiments on the relationship between the Judge and self-consistency are provided in Appendix~\ref{app:exp}.

\section{Conclusion}
We propose an unsupervised self-evolution training framework for multimodal large models. 
By jointly modeling multiple reasoning trajectories from the same input, our method leverages the Actor’s self-consistency signal and a Judge-based modulation.
It further applies group-wise distributional reward modeling to reduce mode collapse during long-term training.
Experiments on multiple mathematical reasoning benchmarks show that our approach achieves stable 
performance improvements.

\section*{\centering Limitations}
However, this work primarily investigates the construction of stable training signals, and leaves the question of how to further improve the self-evolving system beyond the Judge’s capability limit to future study.
To enable sustained unsupervised self-evolution, the Judge should be able to progressively raise its evaluation standards as training proceeds, and autonomously determine when it should be updated to remain a reliable training signal.
\bibliography{custom}

\appendix

\section{Why Group-wise Distributional Modeling Prevents Policy Collapse}
\label{sec:appendix1}
For each input $x$, we sample a set of candidate trajectories $\mathcal{T}(x)=\{\tau_1,\ldots,\tau_n\}$ from the behavior policy $\pi_{\theta_{\mathrm{old}}}(\cdot\mid x)$.  
Each trajectory $\tau_k$ is assigned a final scalar reward $r_k \equiv R(\tau_k,x)$, and we apply an energy scaling $\tilde r_k=\alpha r_k$, where $\alpha$ is a temperature parameter.  
We then introduce a group-wise log-sum-exp baseline
\[
b(x)=\log\sum_{j=1}^{n}\exp(\tilde r_j),
\]
and define the group-relative advantage
\[
A_k(x)=\tilde r_k-b(x).
\]
This construction implicitly induces a target distribution over the candidate set $\mathcal{T}(x)$:
\[
q_\alpha(\tau_k\mid x)\triangleq\frac{\exp(\alpha r_k)}{\sum_{j=1}^{n}\exp(\alpha r_j)}.
\tag{A.1}
\]
It follows immediately that
\[
\log q_\alpha(\tau_k\mid x)
=\alpha r_k-\log\sum_{j=1}^{n}\exp(\alpha r_j)
=A_k(x),
\tag{A.2}
\]
which shows that the group-relative advantage equals the log-probability of $\tau_k$ under the reward-induced distribution $q_\alpha(\cdot\mid x)$.

To clarify the learning objective implied by this distributional modeling, we consider the case where clipping is ignored, and we temporarily omit the KL regularization term.  
In this setting, the dominant gradient term can be written as a policy-gradient form under samples from the behavior policy:
\[
\nabla_\theta \mathcal{J}(\theta)
\propto
\mathbb{E}_{x\sim\mathcal{D},\,\tau\sim\pi_{\theta_{\mathrm{old}}}(\cdot\mid x)}
\big[A(\tau,x)\,\nabla_\theta \log \pi_\theta(\tau\mid x)\big].
\tag{A.3}
\]
Substituting Eq.~(A.2), we obtain
\[
\nabla_\theta \mathcal{J}(\theta)
\propto
\mathbb{E}_{x,\,\tau\sim\pi_{\theta_{\mathrm{old}}}}
\big[\log q_\alpha(\tau\mid x)\,\nabla_\theta \log \pi_\theta(\tau\mid x)\big].
\tag{A.4}
\]
This form suggests that the update is naturally described by matching the policy to the target distribution $q_\alpha(\cdot\mid x)$ defined on the candidate set.

Concretely, consider the following distribution-matching objective over $\mathcal{T}(x)$:
\[
\max_{\theta}\;
\mathbb{E}_{x\sim\mathcal{D}}
\Bigg[
\sum_{k=1}^{n}
q_\alpha(\tau_k\mid x)\,\log \pi_\theta(\tau_k\mid x)
\Bigg].
\tag{A.5}
\]
This objective is equivalent to minimizing the KL divergence on the candidate set:
\[
\min_{\theta}\;
\mathbb{E}_{x\sim\mathcal{D}}
\Big[
D_{\mathrm{KL}}\big(q_\alpha(\cdot\mid x)\,\|\,\pi_\theta(\cdot\mid x)\big)
\Big],
\tag{A.6}
\]
since for any fixed $x$,
\[
D_{\mathrm{KL}}(q\|\pi)
=\sum_k q_k\log q_k
-\sum_k q_k\log \pi_k.
\tag{A.7}
\]
Therefore, maximizing Eq.~(A.5) is equivalent to minimizing Eq.~(A.6).  
By the basic property of KL divergence, the optimum of Eq.~(A.6) satisfies
\[
\pi_\theta(\cdot\mid x)=q_\alpha(\cdot\mid x)\qquad \text{on }\mathcal{T}(x).
\tag{A.8}
\]
This shows that distributional modeling changes the learning target from selecting a single candidate to matching a soft distribution over the candidate set, and the optimal policy approaches the reward-induced distribution $q_\alpha(\cdot\mid x)$.

The sharpness of $q_\alpha(\cdot\mid x)$ is controlled by the temperature $\alpha$ and the reward gaps within the group.  
When $\alpha\to\infty$, or when one candidate has a much larger reward than the others, i.e., $r_{k^\star}\gg r_j$ for all $j\neq k^\star$, the target distribution degenerates to:
\[
q_\alpha(\tau_{k^\star}\mid x)\to 1,\qquad q_\alpha(\tau_j\mid x)\to 0\ (j\neq k^\star).
\tag{A.9}
\]
In this case, the optimal policy in Eq.~(A.8) also becomes deterministic.  
In contrast, when multiple candidates have comparable rewards and no single trajectory dominates, $q_\alpha(\cdot\mid x)$ remains non-degenerate and assigns non-zero probability mass to several candidates.  
Eq.~(A.8) then implies that learning favors a gradual reallocation of probability mass within each group, rather than an immediate collapse to a single mode.

For comparison, a one-hot target distribution over the candidate set,
\[
q_{\mathrm{OH}}(\tau_k\mid x)=\mathbb{I}[k=k^\star],
\tag{A.10}
\]
leads to an optimal solution $\pi_\theta(\tau_{k^\star}\mid x)=1$ when minimizing $D_{\mathrm{KL}}(q_{\mathrm{OH}}\|\pi_\theta)$.  
This objective directly encourages a deterministic mapping.  
By instead using the energy-normalized distribution $q_\alpha(\cdot\mid x)$, the learning target remains a soft distribution as long as $q_\alpha$ is non-degenerate.  
As a result, policy updates can be interpreted as continuously reshaping probability mass within each group, rather than concentrating all mass on a single candidate in early training.

\section{Training Algorithm}
This algorithm~\ref{alg:ours_training} presents the pseudocode of our unsupervised self-evolution training algorithm.
The algorithm summarizes the overall training procedure, including multi trajectory sampling, self consistency based reward initialization, Judge based score modulation, group wise distributional reward shaping, and GRPO based policy optimization.
It is provided for completeness and to facilitate reproducibility.

\begin{algorithm}[t]
\caption{Unsupervised Self-Evolution with Actor--Judge Joint Modeling}
\label{alg:ours_training}

\textbf{Input:} Current policy $\pi_{\theta}$, old policy $\pi_{\theta_{\mathrm{old}}}$, unlabeled training set $\mathcal{D}$,
group size $G$, frozen Judge $J_{\phi}$, reference model $\pi_{\mathrm{ref}}$,
energy temperature $\alpha$, clip parameter $\epsilon$, KL coefficient $\beta$,
answer extractor $E(\cdot)$, calibration params $(\lambda_{+},\lambda_{-},t_h,t_l,\tau_h,\tau_l)$.

\ForEach{$(I,q)\sim \mathcal{D}$}{
    Sample a group of trajectories: $\{\tau_i\}_{i=1}^{G} \sim \pi_{\theta_{\mathrm{old}}}(\cdot\mid I,q)$\;
    \tcp*[r]{\algcmt{// Sample multiple trajectories}}

    Extract answers: $\hat{Y}=E(\mathcal{T})=\{\hat{y}_i\}_{i=1}^{G}$\;
    \tcp*[r]{\algcmt{// Parse final answers}}

    Compute self-consistency distribution:
    $\hat{p}(y)=\frac{1}{G}\sum_{i=1}^{G}\mathbb{I}[\hat{y}_i=y]$\;
    Assign SC rewards: $r_i^{\mathrm{SC}} \leftarrow \hat{p}(\hat{y}_i)$\;
    \tcp*[r]{\algcmt{// Soft frequency reward within the group}}

    Judge scoring (frozen): $s_i \leftarrow J_{\phi}(I,q,\tau_i)\in[0,1]$\;
    \tcp*[r]{\algcmt{// Trajectory-level evaluation}}

    Calibrate Judge scores:
    $g(s_i) \leftarrow 1+\lambda_{+}\sigma\!\left(\frac{s_i-t_h}{\tau_h}\right)-\lambda_{-}\sigma\!\left(\frac{t_l-s_i}{\tau_l}\right)$\;
    \tcp*[r]{\algcmt{// Bounded, smooth modulation}}

    Compute final rewards: $R_i \leftarrow r_i^{\mathrm{SC}}\cdot g(s_i)$\;
    \tcp*[r]{\algcmt{// Joint modeling: SC $\times$ Judge modulation}}

    Group-wise distributional shaping:
    $\tilde{r}_i \leftarrow \alpha R_i$;\quad
    $b \leftarrow \log\sum_{j=1}^{G}\exp(\tilde{r}_j)$;\quad
    $A_i \leftarrow \tilde{r}_i-b$\;
    \tcp*[r]{\algcmt{// Group-relative advantage via log-sum-exp baseline}}

    Compute GRPO objective $\mathcal{J}(\theta)$ according to Eq.~(12) , Eq.~(13) and Eq.~(14)\;
    \tcp*[r]{\algcmt{// Group-relative policy optimization}}
    where $\gamma_{i,t}(\theta)=\frac{\pi_{\theta}(o_{i,t}\mid I,q,o_{i,<t})}{\pi_{\theta_{\mathrm{old}}}(o_{i,t}\mid I,q,o_{i,<t})}$\;
    \tcp*[r]{\algcmt{// Token-level ratio as in GRPO}}

    Update policy parameters: $\theta \leftarrow \theta + \eta\,\nabla_{\theta}\mathcal{J}(\theta)$\;
    Update old policy: $\theta_{\mathrm{old}} \leftarrow \theta$\;
}
\textbf{return} $\pi_{\theta}$\;
\end{algorithm}

\section{Experimental Setup and Baselines}
\subsection{Training Data}
Geo3K\cite{geo3k} is a geometry problem-solving dataset designed to support diagram-grounded symbolic reasoning. 
It contains 3,002 multiple-choice geometry questions, each paired with a natural language problem statement, a corresponding geometry diagram, and a ground-truth answer.
Geo3K is split into 2,101 / 300 / 601 examples for training, validation, and testing, respectively. 
The original paper also reports basic statistics for each split, such as the numbers of sentences and words, as well as the scale of formal semantic annotations, indicating the dataset’s semantic and reasoning complexity. 

GeoQA\cite{geoqa} is a benchmark for multimodal numerical reasoning in planar geometry.
Each example is a geometry problem collected from real exam or practice question banks, where the model must jointly interpret the problem text and the accompanying diagram.
The original GeoQA dataset contains 5,010 problems and follows a 7:1.5:1.5 split for training, validation, and test sets.
GeoQA further groups problems into three categories: angle computation, length computation, and others (e.g., area-related problems).

The MMR1 training data\cite{mmr1} is organized into two parts: a cold-start set with long chain-of-thought (CoT) annotations and an RL-stage set of question–answer (QA) pairs. 
This design aims to cover both multimodal mathematical and logical reasoning, with an emphasis on data quality, difficulty, and diversity.
In our experiments, we use the RL-stage QA split as the unlabeled training set for unsupervised self-evolution.

\subsection{Baselines}
Vision-Zero\cite{Vision-zero} proposes a zero-human-in-the-loop framework for improving vision–language models (VLMs). Instead of relying on manually curated QA pairs or preference data, it formulates training as a self-play game based on visual differences, inspired by the “Who Is the Spy” setting.
The key idea is that, given a pair of slightly different images (an original image and its edited counterpart), the model can generate training signals through multi-round interactions and optimize with verifiable rewards, thereby improving visual reasoning and strategic inference.
In our comparisons, we use two Vision-Zero variants built on Qwen2.5-VL-7B: one trained with image pairs constructed from synthetic CLEVR scenes, and the other trained with real-world edited image pairs from datasets such as ImgEdit.

EvoLMM\cite{EvoLMM} trains using only raw images, without any human-annotated QA pairs or metadata.
It improves multimodal reasoning through a closed-loop process where the model generates questions, produces answers, and derives rewards from its own outputs.
To build the training pool, EvoLMM samples about 1,000 images from each of several widely used visual reasoning, chart, and geometry datasets, resulting in roughly 6k training images.
The method adopts a Proposer–Solver framework, where the Proposer generates questions and the Solver answers them to drive self-evolution.

\subsection{Evaluation Benchmarks}
MathVision\cite{mathvision} is a benchmark designed to evaluate large models’ mathematical reasoning under visual context.
The authors argue that existing visual math benchmarks, while broad, are still limited in problem diversity and subject coverage; 
MathVision is therefore introduced to provide a more comprehensive assessment of vision-based mathematical reasoning.
Following common practice, we evaluate both our method and all baselines on the official testmini split, which contains 304 problems.

MathVerse\cite{mathverse} is designed not only to measure final-answer accuracy, but also to diagnose whether multimodal large models truly use diagram information in visual math problems.
The authors note that many existing benchmarks include substantial textual cues that duplicate the visual content, allowing models to answer correctly with little reliance on the image and thus overestimating genuine visual understanding.
MathVerse covers three major areas—Plane Geometry, Solid Geometry, and Functions—and provides 12 fine-grained categories for capability-based evaluation.
In our experiments, we evaluate on the MathVerse-mini split.

WeMath\cite{we-math} is motivated by the observation that most visual math benchmarks focus on final-answer accuracy, but provide limited insight into a model’s underlying weaknesses in mathematical knowledge and generalization. 
It therefore introduces a textbook concept–centered evaluation framework to distinguish whether errors arise from missing knowledge or from failure to compose and generalize known concepts.
WeMath categorizes model behaviors into four diagnostic types: IK (Insufficient Knowledge), IG (Inadequate Generalization), CM (Complete Mastery), and RM (Rote Memorization).
We evaluate both our method and all baselines on the full WeMath benchmark, which contains 6.5K visual math problems.

LogicVista\cite{logicvista} is a benchmark designed to evaluate logical reasoning of multimodal large language models (MLLMs) under visual context. 
The authors argue that existing multimodal evaluations often focus on perception and understanding, or emphasize math-oriented reasoning, while providing limited coverage of more general logical reasoning skills needed for tasks such as navigation and pattern inference.
LogicVista consists of 448 multiple choice questions. It groups logical reasoning into five skill categories: inductive reasoning, deductive reasoning, numerical reasoning, spatial reasoning, and mechanical reasoning.
We evaluate on the full LogicVista benchmark.

DynaMath\cite{Dynamath} focuses on the robustness of mathematical reasoning.
Unlike most existing visual math benchmarks that are static, it is designed to systematically test whether the same reasoning process remains valid under varying conditions. 
To this end, DynaMath evaluates generalization by dynamically generating problem variants.
The benchmark contains 501 high-quality seed questions spanning multiple topics, and each seed is instantiated into 10 variants, resulting in 5,010 instances in total.
We evaluate on the full DynaMath benchmark.

\section{Prompt Design}
This section presents the Judge prompt and the Actor prompt used during training to support reward modeling and output-format constraints in our unsupervised self-evolution framework.

The Judge prompt consists of two complementary components. 
The first component specifies the scoring and decision rules, which evaluate each candidate solution trajectory along three dimensions: answer correctness, reasoning quality, and visual grounding. 
The second component enforces a strict output format to standardize the Judge’s scores and ensure stable reward signals. We use the combination of these two components as the full Judge prompt.

The Actor prompt is not designed to elicit better reasoning content. 
Instead, it enforces a unified output format.
This constraint helps avoid ambiguous formatting, missing answers, or multiple final answers, and prevents the Judge from producing unstable or exploitable scores due to malformed outputs during reward modeling.
Importantly, these prompts are only used in the training stage for self-evolution.
For all evaluations, we use the original, official prompting setup of each model and keep the evaluation protocol identical across all methods.

\begin{tcolorbox}[
  colback=yellow!10!white,
  colframe=yellow!50!black,
  title=\textbf{Judge Prompt},
  fonttitle=\bfseries,
  sharp corners]

\textit{You are an expert evaluator for multimodal mathematical reasoning.}

All scores must be real numbers in $[0,1]$.

\textbf{Evaluation Dimensions:}

\textbf{1. answer\_correctness}
\begin{itemize}
  \setlength{\itemsep}{0pt}
  \setlength{\topsep}{2pt}
  \item 1.0: fully correct
  \item 0.5--0.9: almost correct
  \item 0.1--0.4: partial progress
  \item 0.0: wrong or missing
\end{itemize}

\textbf{2. reasoning\_quality}
\begin{itemize}
  \setlength{\itemsep}{0pt}
  \setlength{\topsep}{2pt}
  \item 1.0: rigorous and logical
  \item 0.5--0.9: minor gaps
  \item 0.1--0.4: fragmented reasoning
  \item 0.0: invalid reasoning
\end{itemize}

\textbf{3. visual\_grounding}
\begin{itemize}
  \setlength{\itemsep}{0pt}
  \setlength{\topsep}{2pt}
  \item 1.0: correct use of visual elements
  \item 0.5--0.9: minor misreadings
  \item 0.1--0.4: partial or incorrect usage
  \item 0.0: no grounding or hallucination
\end{itemize}

\textbf{Overall Score}

\begin{itemize}
  \setlength{\itemsep}{0pt}
  \setlength{\topsep}{2pt}
  \item answer\_correctness: 0.50
  \item reasoning\_quality: 0.30
  \item visual\_grounding: 0.20
\end{itemize}

\textbf{Mandatory Validity Rule}

If the candidate solution fails to provide a clear and single final answer in the required format, \textbf{all scores are set to \texttt{0.0}}.

\end{tcolorbox}

\begin{tcolorbox}[
  colback=gray!5!white,
  colframe=gray!50!black,
  title=\textbf{Judge Output Format},
  fonttitle=\bfseries,
  sharp corners
]

\begin{verbatim}
[BEGIN THOUGHT]
<your analysis>
[END THOUGHT]

[BEGIN SCORES]
{
  "answer_correctness": <float>,
  "reasoning_quality": <float>,
  "visual_grounding": <float>,
  "overall_score": <float>
}
[END SCORES]
\end{verbatim}

Only valid JSON enclosed in the above tags is accepted.

\end{tcolorbox}

\begin{tcolorbox}[
  colback=blue!6!white,
  colframe=blue!50!black,
  title=\textbf{Actor Prompt},
  fonttitle=\bfseries,
  sharp corners
]

You must first think through the reasoning process as an internal monologue.
The entire reasoning process must be fully enclosed within a matching
\texttt{<think>...</think>} pair, with no text outside these tags.

After the closing \texttt{</think>} tag, you must output the final answer
in the exact format \texttt{\textbackslash boxed\{ANSWER\}}.

The \texttt{\textbackslash boxed\{\}} must contain only the final answer value,
with no additional words, steps, symbols, units, or punctuation.

If the required format is not followed \textbf{exactly}, the answer is
considered invalid.

\end{tcolorbox}

\section{Hyperparameters}
This section summarizes the main hyperparameters and system configurations used in our unsupervised self-evolution training, to support reproducibility and clarify implementation details.Detailed settings are provided in Table~\ref{tab:hyperparams}.

\begin{table}[t]
\centering
\caption{Training hyperparameters.}
\label{tab:hyperparams}
\renewcommand{\arraystretch}{1.12}
\setlength{\tabcolsep}{6pt}
\small
\begin{tabular}{p{0.55\linewidth}p{0.35\linewidth}}
\toprule
\textbf{Category} & \textbf{Hyperparameter (Value)} \\
\midrule

\rowcolor{gray!12}
\multicolumn{2}{l}{\textbf{Data Configuration}} \\
Train Batch Size & 512 \\
Max Prompt Length & 1024 \\
Max Response Length & 2048 \\
Filter Overlong Prompts & True \\
Truncation Strategy & \texttt{error} \\
Image Key & \texttt{images} \\
\midrule

\rowcolor{gray!12}
\multicolumn{2}{l}{\textbf{Model \& Optimization}} \\
Base Model & Qwen2.5-VL-7B-Instruct \\
Optimizer Learning Rate & $1\times 10^{-6}$ \\
KL Loss Coefficient ($\beta$) & 0.01 \\
KL Loss Type & Low-Var KL \\
\midrule

\rowcolor{gray!12}
\multicolumn{2}{l}{\textbf{PPO / GRPO Settings}} \\
Algorithm & GRPO \\
PPO Mini Batch Size & 128 \\
PPO Micro Batch Size (per GPU) & 8 \\
Rollout Group Size ($n$) & 8 \\
Ref Log-Prob Micro Batch (per GPU) & 20 \\
Param Offload (Ref Model) & True \\
\midrule

\rowcolor{gray!12}
\multicolumn{2}{l}{\textbf{Sampling / Engine}} \\
Rollout Engine & \texttt{\$ENGINE} \\
Tensor Model Parallel Size & 2 \\
GPU Memory Utilization (Rollout) & 0.7 \\
GPU Memory Utilization (Reward) & 0.6 \\
\midrule

\rowcolor{gray!12}
\multicolumn{2}{l}{\textbf{Trainer Settings}} \\
Total Epochs & 20 \\
GPUs per Node & 8 \\
Number of Nodes & 1 \\
\bottomrule
\end{tabular}
\end{table}

\section{Case study}
\begin{figure*}[t]
    \centering
    \includegraphics[width=\textwidth]{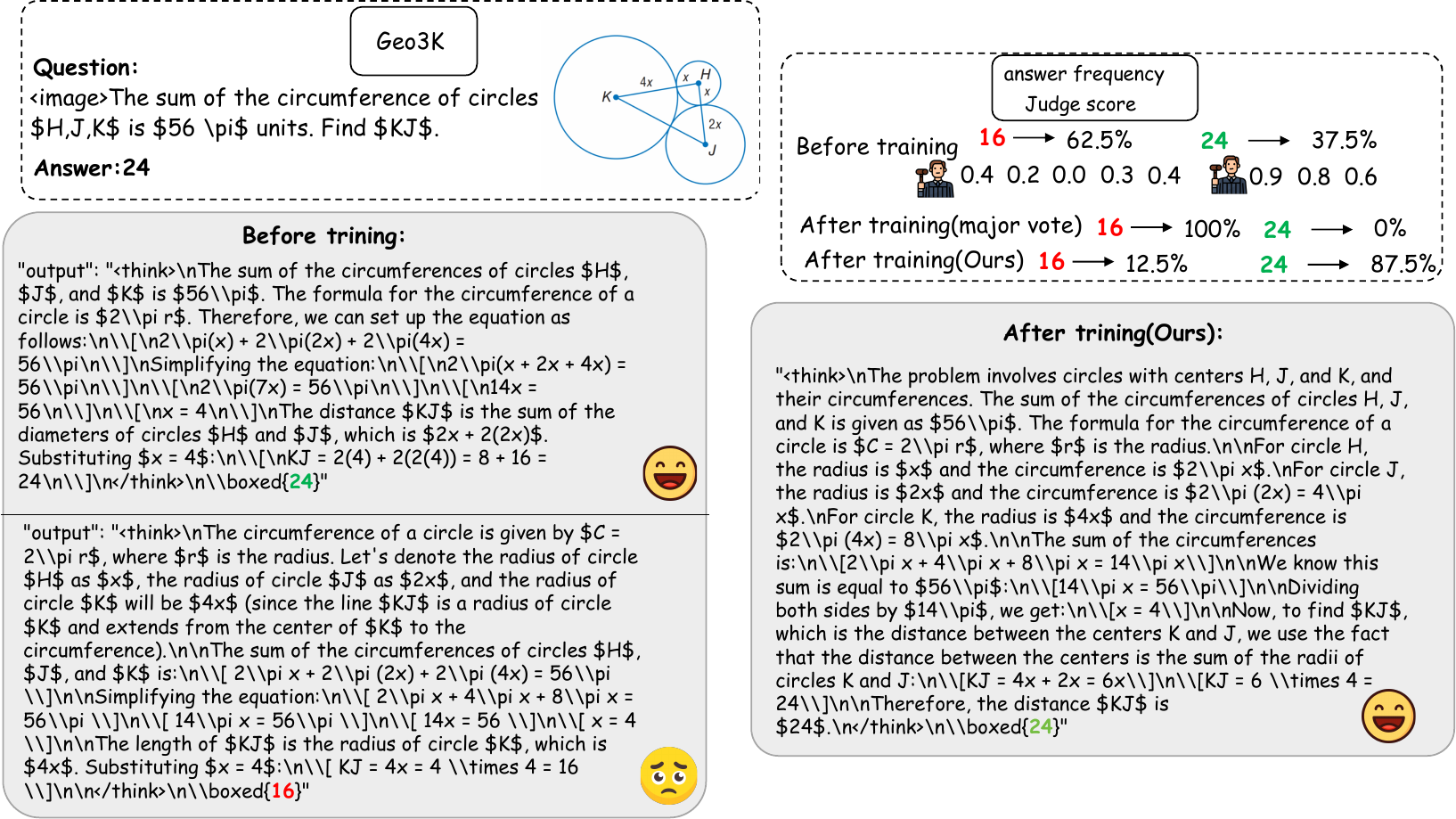}
    \caption{A case study on Geo3K\cite{geo3k}}
    \label{case1}
\end{figure*}

\begin{figure*}[t]
    \centering
    \includegraphics[width=\textwidth]{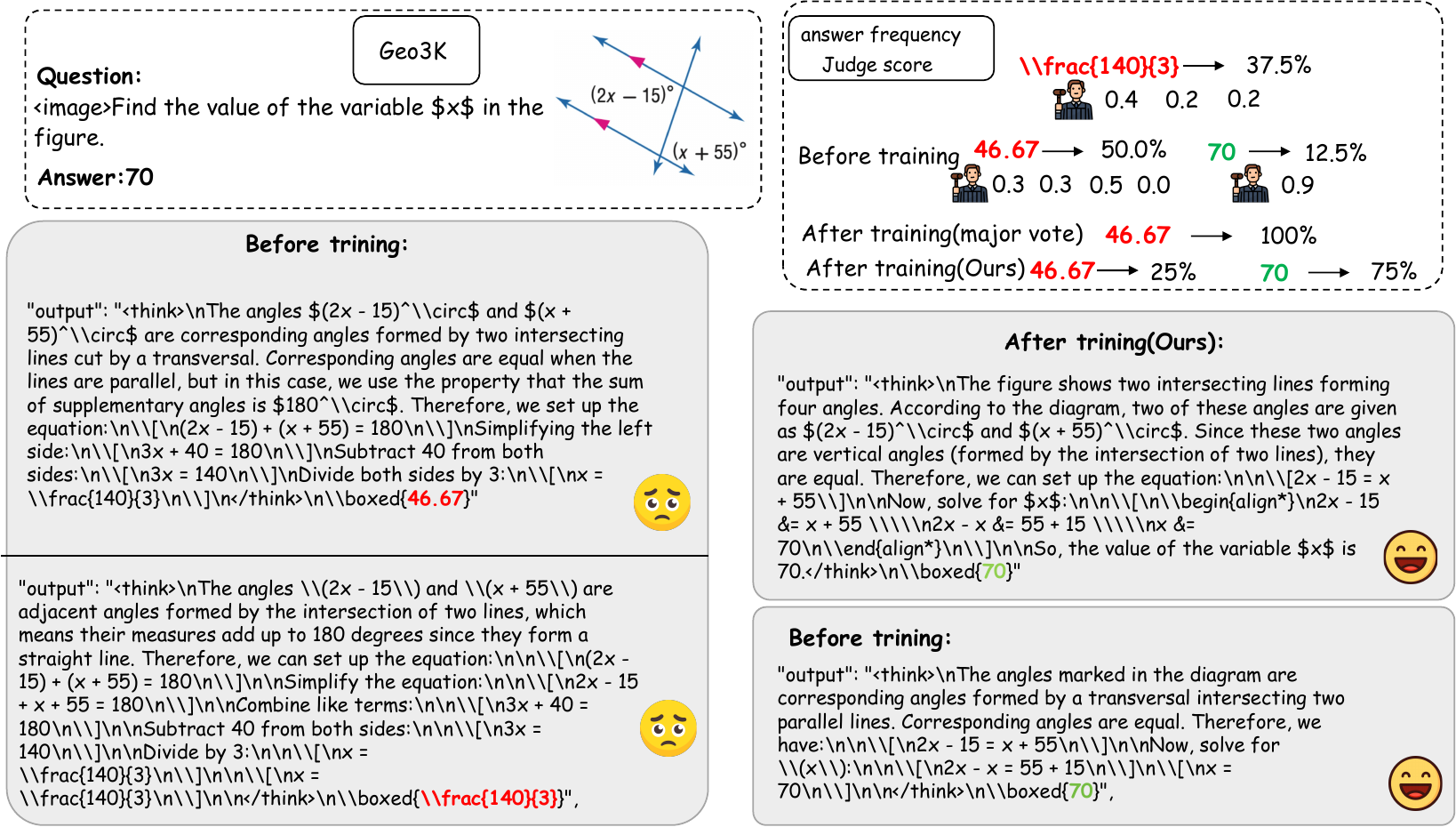}
    \caption{A case study on Geo3K\cite{geo3k}}
    \label{case2}
\end{figure*}

In this section, we present qualitative case studies to illustrate the behavior of our method and analyze representative failure cases. 
Through these examples, we provide deeper insights into how the proposed framework influences model reasoning and where its limitations remain.
Figures~\ref{case1} and ~\ref{case2} compare the behaviors of our method and the majority-voting strategy during training. 
Majority voting essentially compresses the model’s output distribution into a near-deterministic mapping: the model updates only with respect to the most frequent answer, while ignoring the relative structure among alternative candidates for the same question. 
In the early stage of training, even a slight frequency advantage of an incorrect answer over the correct one can be amplified over subsequent iterations. 
As the incorrect answer becomes dominant, its pseudo-label signal is repeatedly reinforced, eventually driving the model into a self-reinforcing but incorrect learning trajectory.
In contrast, our method exhibits more stable learning behavior overall.

\begin{figure*}
    \centering
    \begin{subfigure}[t]{\textwidth}
        \centering
        \includegraphics[width=\textwidth]{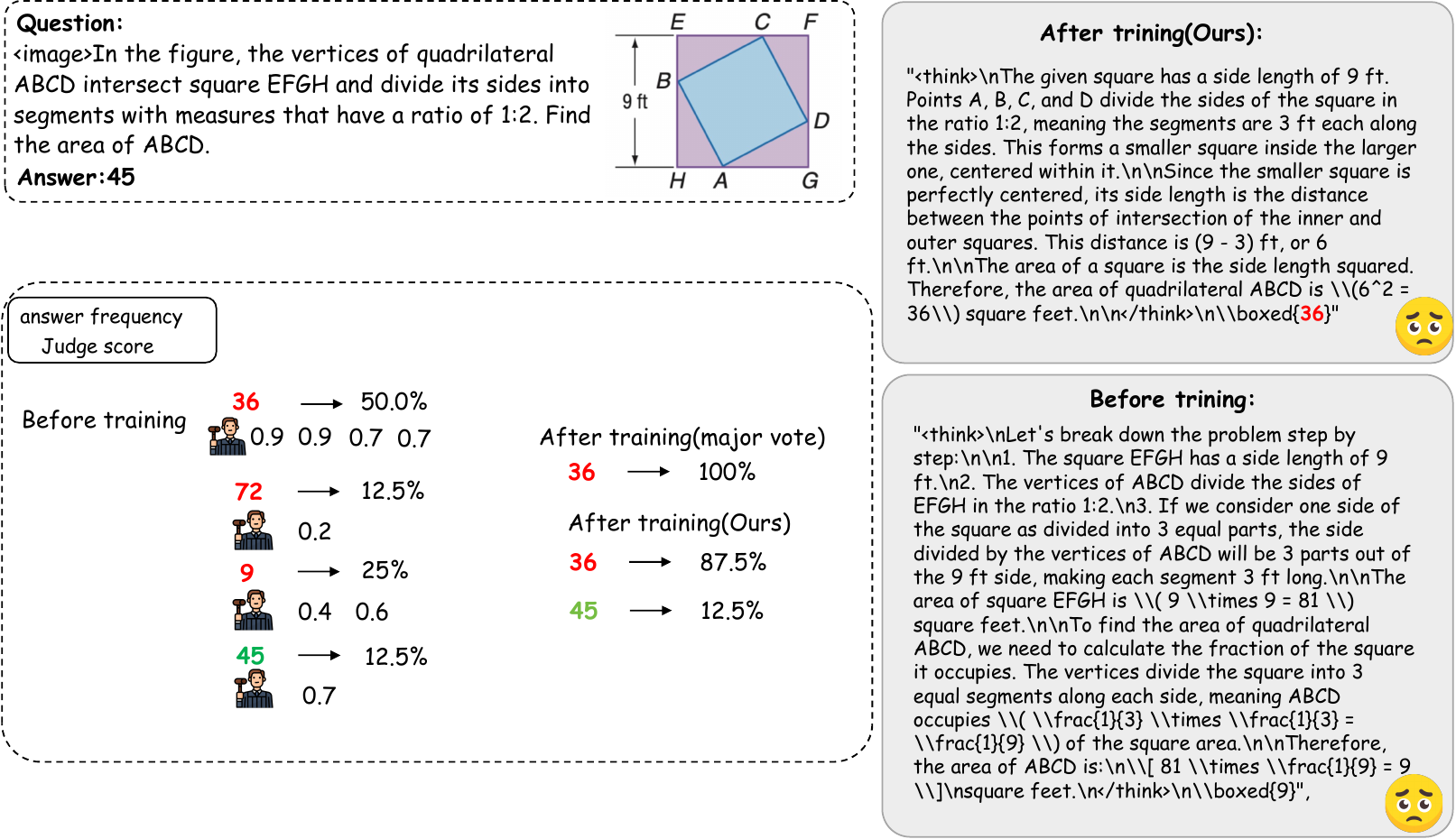}
        \label{case3:mv}
    \end{subfigure}

    \vspace{0.5em}

    \begin{subfigure}[t]{\textwidth}
        \centering
        \includegraphics[width=\textwidth]{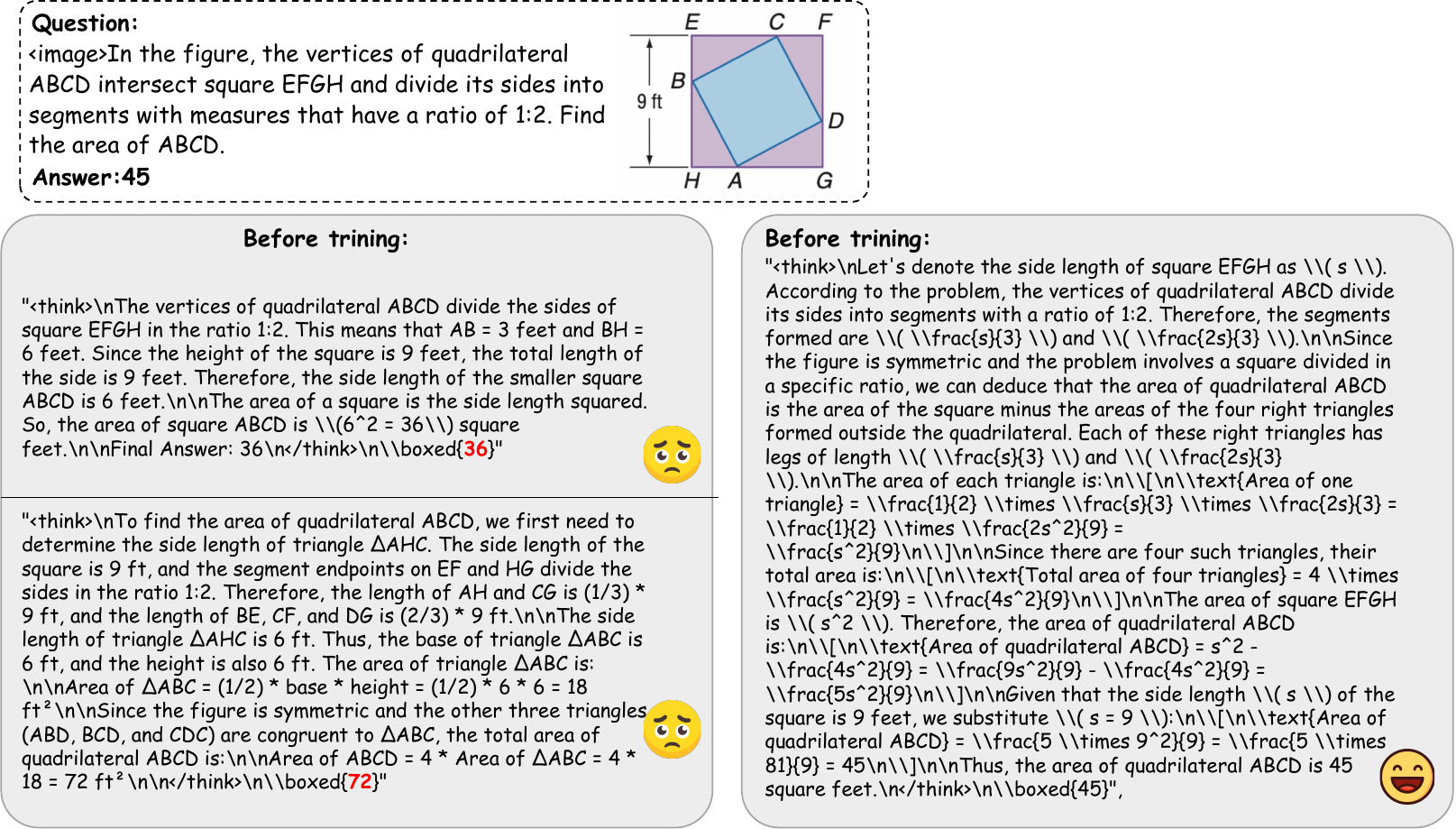}
        \label{case3:ours}
    \end{subfigure}
    \caption{A qualitative failure case analysis on the Geo3K dataset~\cite{geo3k}.}
    \label{case3}
\end{figure*}

Figure~\ref{case3} presents a representative failure case. 
In some cases, the training signal can still be misled when the model’s self-consistency distribution is already strongly biased toward an incorrect answer and the frozen Judge also assigns a relatively high score to that answer. 
Under this “incorrect consensus” scenario, the group-relative reward modeling will continue to favor the wrong trajectory, causing the model to update in an undesired direction. 
Moreover, once such incorrect consistency is solidified, the output distribution becomes sharper and exploration is reduced. 
This effect can partially explain the drop in pass@10 reported in Table~\ref{tab4}: with lower sampling diversity, the probability of reaching the correct solution across multiple attempts decreases.

\section{Experiments and Analysis}
\label{app:exp}

\paragraph{\textbf{Analysis of pass@10.}}
Table~\ref{tab4} reports pass@10 results of different methods across multiple benchmarks. 
Note that pass@10 is highly sensitive to sampling diversity, as it measures the probability of obtaining at least one correct solution among multiple samples. 
Our method improves training stability through group-relative reward modeling and Judge-based modulation, reducing the random amplification of incorrect trajectories, but it can also make the output distribution more concentrated. 
In some cases, especially when both the model’s self-consistency signal and the Judge score favor the same incorrect answer, the policy may develop an incorrect consensus, which further reduces sampling diversity. 
Such distributional contraction weakens the coverage benefit of multiple attempts and is one of the main reasons for the decrease in pass@10.

\paragraph{\textbf{Effect of Distributional Modeling.}}
Table~\ref{tab:abl6} analyzes the effect of group-wise distributional modeling under different designs. 
Across self-consistency, Judge-only, and our final formulation, incorporating distributional modeling consistently leads to improved performance on both benchmarks. 
This indicates that distributional modeling serves as a general and stable refinement step, converting raw trajectory-level signals into relative group-wise advantages, which helps smooth the training signal and leads to more stable optimization.

\paragraph{\textbf{Relationship Between Self-Consistency and Judge Preferences.}}
The results are presented in Table~\ref{tab:abl7}.
Agree@1 measures the fraction of questions for which the top-1 choice under self-consistency (i.e., the answer with the highest self-consistency) matches the top-1 answer selected by the Judge (i.e., the answer with the highest average score), reflecting the alignment between the two signals over the candidate set. 
SC-winner denotes the answer accuracy when selecting solely based on self-consistency, serving as a proxy for the reliability of self-consistency as a distributional prior. 
J-winner denotes the answer accuracy when selecting solely based on the Judge. 
For evaluation, we sample a batch from Geo3K and, for each input, roll out 8 trajectories using both the base model and the final model.

As shown in Table~\ref{tab:abl7}, the top-1 agreement between self-consistency and the Judge increases noticeably after training while remaining unsaturated.
This suggests that the model’s output distribution gradually shifts from being partially driven by incidental frequency patterns to being more aligned with reasoning quality. 
At the same time, the agreement remaining below saturation indicates that the policy does not collapse into a deterministic mapping dominated by a single signal, and still preserves room for exploration.

Moreover, the accuracy of the trajectories selected by the Judge improves, even though the Judge is not trained.
This implies that the Judge guidance encourages the Actor to produce higher quality reasoning trajectories more consistently. 
Overall, these results highlight the synergy between self-consistency and Judge-based modulation: self-consistency provides a stable distributional prior, while the Judge steers learning toward higher quality candidates under ambiguity. 
They form a positive feedback loop that improves trajectory quality while mitigating distribution collapse.
We present the full evaluation results of unsupervised training on Geo3K across multiple benchmarks in Tables~\ref{tab:geo3k},~\ref{tab:mathverse_category},~\ref{tab:logicvista_category},~\ref{tab:wemath_selected}.

\section{Related work}
Recent advances show that reinforcement learning (RL) enhances the reasoning ability of large language models (LLMs), with representative systems including DeepSeek-R1\cite{DeepSeek-R1} and Kimi-K1.5\cite{Kimi-K1.5}.
Under appropriately designed optimization objectives, models can gradually acquire more long-term strategic reasoning behaviors.
These advances are largely enabled by effective RL-based optimization frameworks, including PPO\cite{PPO}, DPO\cite{DPO}, GRPO\cite{GRPO}, DAPO\cite{DAPO}, and GSPO\cite{GSPO}.  
\subsection{Multi-modal Reasoning}
Motivated by the success of verifiable rewards in LLM reasoning, recent studies\cite{VLM-R1} have begun to explore post-training and R1-style reinforcement learning in multimodal settings.
Instead of relying on subjective human preferences, these methods\cite{R1-Onevision,Vision-R1} derive reward signals from objectively verifiable signals, enabling more stable reasoning optimization.
Empirical results show that when rewards are verifiable or targets are well structured, RL-style post-training leads to stable improvements on multimodal reasoning tasks.

To address this limitation, another line of research\cite{MMC} explicitly introduces reflection mechanisms to improve robustness during reasoning.
For example, VL-Rethinker\cite{VL-Rethinker} studies self-reflection in multimodal reasoning and examines the trade-off between reasoning benefits and computational cost.
Building on this idea, later work\cite{R3V,LLAVA-Critic-R1} integrates reflection into training by using structured reflection steps or learning an explicit critic for evaluation.
NaturalReasoning\cite{Yuan2025NaturalReasoningRI} proposes a method for constructing large-scale reasoning data from real-world corpora. 
It shows that such large-scale in-the-wild reasoning data can support both knowledge distillation from strong teacher models and self-training with either external reward models or self-generated rewards.
Building on this line of work, NaturalThoughts\cite{Li2025NaturalThoughtsSA} studies which teacher-generated reasoning traces are the most useful for distillation. 
Based on the large-scale question set introduced in NaturalReasoning, it selects high-quality reasoning traces generated by strong teacher models for distillation training.
R2-MultiOmnia\cite{ranaldi-etal-2025-r2} presents a self-training framework for multilingual multimodal reasoning. 
Despite these advances, effective reasoning post-training still relies on high-quality training signals or stronger teacher models.
\subsection{Self-Evolving In Large Language Models}
Unsupervised self-evolution has been explored to some extent in large language models\cite{CAN_LARGE_REASONING_MODELS_SELF-TRAIN?}.
A core idea is that, even without ground-truth answers, test-time scaling strategies (e.g., majority voting) can provide useful relative correctness signals\cite{TTRL,ETTRL}.
In parallel, some work\cite{No_Free_Lunch,Learning_to_Reason_without_External_Rewards} uses reinforcement learning with internal feedback, treating model-internal signals (e.g., confidence) as rewards and eliminating the need for external annotations.
Going further, some methods\cite{Self-Questioning_Language_Models} move beyond a fixed training set by letting models generate tasks and improve themselves.
Self-Empowering VLMs\cite{Yang2025SelfEmpoweringVA} studies hierarchical understanding in VLMs and shows that the main challenge is not missing taxonomic knowledge, but the difficulty of maintaining cross-level consistency during step-by-step prediction.
Absolute Zero\cite{Absolute_Zero} studies a fully data-free setting where the model generates tasks and uses executable checkers to verify answers, providing a self-driven curriculum and RLVR signals.
Building on this idea, R-Zero\cite{R-Zero} uses a Challenger–Solver co-evolution framework to generate suitable problems.

Recently, self-evolution has also been extended to multimodal large language models.
MM-UPT\cite{MM-UPT} uses majority voting over multiple sampled answers to form pseudo-rewards, enabling continual improvement on multimodal reasoning data without ground-truth labels.
EvoLMM\cite{EvoLMM} uses a Proposer–Solver loop and derives continuous self-rewards from internal consistency signals.
VisPlay\cite{VisPlay} uses a Questioner–Reasoner role split and applies GRPO with diversity rewards to balance question complexity and answer quality, enabling autonomous evolution from unlabeled images.
However, most of these methods use majority voting as the main training signal, which primarily reinforces consistency under the current output distribution. 
Over long-term training, this can bias the model toward early dominant patterns and limit exploration.

\begin{table}[t]
\centering
\small
\begin{tabular}{lcc}
\toprule
Method & MathVision & DynaMath \\
\midrule
Self-Consistency        & 25.2 & 20.5 \\
w.\ distribution        & 25.9 & 20.7 \\
\midrule
Judge                   & 27.3 & 21.1 \\
w.\ distribution        & 27.8 & 21.5 \\
\midrule
Ours w.o.\ distribution & 30.1 & 23.7 \\
Ours w.\ distribution   & 30.9 & 24.2 \\
\bottomrule
\end{tabular}
\caption{Effect of group-wise distributional modeling.}
\label{tab:abl6}
\end{table}

\begin{table}[t]
\centering
\small
\begin{tabular}{lccc}
\toprule
Method & Agree@1 & SC-winner & J-winner \\
\midrule
Qwen2.5-VL-7B & 41.2 & 36.3 & 40.2 \\
Ours          & 73.8 & 48.6 & 50.3 \\
\bottomrule
\end{tabular}
\caption{Relationship between self-consistency and Judge preferences.}
\label{tab:abl7}
\end{table}

\begin{table*}[t]
\centering
\small
\setlength{\tabcolsep}{4pt}
\renewcommand{\arraystretch}{1.12}

\caption{Category-wise overall accuracy (\%) on MathVision (trained on Geo3K).}
\label{tab:geo3k}

\begin{tabular}{lccccc}
\toprule
\textbf{Method} & \textbf{Overall} & \textbf{Algebra} & \textbf{Analytic Geo.} & \textbf{Arithmetic} & \textbf{Comb. Geo.} \\
\midrule
Qwen2.5-VL-7B & 25.0 & 15.8 & 26.3 & 36.8 & 15.8 \\
Ours          & 30.9 & 15.8 & 42.1 & 36.8 & 21.1 \\
\midrule
\textbf{Method} & \textbf{Combinatorics} & \textbf{Counting} & \textbf{Descriptive Geo.} & \textbf{Graph Theory} & \textbf{Logic} \\
\midrule
Qwen2.5-VL-7B & 10.5 & 21.1 & 26.3 & 15.8 & 15.8 \\
Ours          & 21.1 & 21.1 & 21.1 & 10.5 & 21.1 \\
\midrule
\textbf{Method} & \textbf{Metric (Angle)} & \textbf{Metric (Area)} & \textbf{Metric (Length)} & \textbf{Solid Geo.} & \textbf{Statistics} \\
\midrule
Qwen2.5-VL-7B & 36.8 & 47.4 & 31.6 & 15.8 & 47.4 \\
Ours          & 63.2 & 47.4 & 21.1 & 36.8 & 57.9 \\
\midrule
\textbf{Method} & \textbf{Topology} & \textbf{Transformation Geo.} & \multicolumn{3}{c}{ } \\
\midrule
Qwen2.5-VL-7B & 26.3 & 10.5 & \multicolumn{3}{c}{ } \\
Ours          & 36.8 & 21.1 & \multicolumn{3}{c}{ } \\
\bottomrule
\end{tabular}
\end{table*}

\begin{table*}[t]
\centering
\small
\setlength{\tabcolsep}{6pt}
\renewcommand{\arraystretch}{1.15}

\caption{Category-wise overall accuracy (\%) on MathVerse (trained on Geo3K).}
\label{tab:mathverse_category}

\begin{tabular}{lccccc}
\toprule
Method & Text Dominant & Vision Only & Vision Dominant & Vision Intensive & Text Lite \\
\midrule
Qwen2.5-VL-7B & 53.6 & 41.0 & 40.9 & 40.9 & 44.9 \\
Ours          & 56.6 & 42.5 & 42.6 & 44.8 & 47.7 \\
\bottomrule
\end{tabular}
\end{table*}

\begin{table*}[t]
\centering
\small
\setlength{\tabcolsep}{8pt}
\renewcommand{\arraystretch}{1.15}

\caption{Category-wise overall accuracy (\%) on LogicVista(trained on Geo3K).}
\label{tab:logicvista_category}

\begin{tabular}{lcccccc}
\toprule
Method & Overall & Inductive & Deductive & Numerical & Spatial & Mechanical \\
\midrule
Qwen2.5-VL-7B & 46.3 & 36.4 & 64.5 & 55.8 & 19.2 & 54.1 \\
Ours          & 49.0 & 29.0 & 65.6 & 65.3 & 30.8 & 55.4 \\
\bottomrule
\end{tabular}
\end{table*}

\begin{table*}[t]
\centering
\small
\setlength{\tabcolsep}{7pt}
\renewcommand{\arraystretch}{1.15}

\caption{Category-wise overall accuracy (\%) on Wemath (trained on Geo3K).}
\label{tab:wemath_selected}

\begin{tabular}{lccc}
\toprule
Category & Qwen2.5-VL-7B & Ours & $\Delta$ \\
\midrule
Two-step (S2)                          & 55.28 & 58.61 & +3.33 \\
Understanding of Plane Figures         & 60.22 & 64.45 & +4.23 \\
Calculation of Solid Figures           & 77.26 & 79.03 & +1.77 \\
Direction                              & 82.86 & 90.00 & +7.14 \\
Route Map                              & 55.49 & 62.64 & +7.15 \\
\bottomrule
\end{tabular}
\end{table*}

\end{document}